\documentclass{article}
\usepackage{spconf,amsmath,graphicx}
\usepackage{amsmath}
\usepackage{amssymb}

\title{Meta-3DSeg: Learning to learn 3D Shape Segmentation Functions}
\name{Yu Hao, Hao Huang, Shuaihang Yuan, Yi Fang}
\address{NYU Multimedia and Visual Computing Lab\\
New York University Abu Dhabi\\
Abu Dhabi, UAE\\}
\begin{document}
%
\maketitle
\begin{abstract}
Learning robust 3D shape segmentation functions with deep neural networks has emerged as a powerful paradigm, offering promising performance in producing a consistent part segmentation of each 3D shape. Generalizing across 3D shape segmentation functions requires robust learning of priors over the respective function space and enables consistent part segmentation of shapes in presence of significant 3D structure variations. Existing generalization methods rely on extensive training of 3D shape segmentation functions on large-scale labeled datasets. In this paper, we proposed to formalize the learning of a 3D shape segmentation function space as a meta-learning problem, aiming to predict a 3D segmentation model that can be quickly adapted to new shapes with no or limited training data. More specifically, we define each task as unsupervised learning of shape-conditioned 3D segmentation function which takes as input points in 3D space and predicts the part-segment labels. The 3D segmentation function is trained by a self-supervised 3D shape reconstruction loss without the need for part labels. Also, we introduce an auxiliary deep neural network as a meta-learner which takes as input a 3D shape and predicts the prior over the respective 3D segmentation function space. We show in experiments that our meta-learning approach, denoted as Meta-3DSeg, leads to improvements on unsupervised 3D shape segmentation over the conventional designs of deep neural networks for 3D shape segmentation functions.
\end{abstract}
\begin{keywords}
3D segmentation, meta-learning
\end{keywords}
\section{Introduction}
3D part segmentation is a meaningful but challenging task, which has wide applications in the 3D computer vision field \cite{sidi2011unsupervised,wu2013unsupervised,wang2020few}. This task requires us to classify each sampled point from the target 3D shape into meaningful part clusters. As large-scale 3D shapes datasets become available to the public \cite{chang2015shapenet}, learning-based data-driven methods become popular and demonstrate great success in part segmentation of 3D shapes \cite{qi2017pointnet,qi2017pointnet++}. Earlier supervised learning-based models leverage deep neural networks as 3D shape segmentation functions which firstly encode every 3D point to a high-dimensional point signature, and then map the signature to the corresponding semantic part label. The deep learning models are usually trained with a large-scale labeled dataset to optimize the parameters of a model towards robust 3D segmentation. Recent advances show that neural implicit shape representation parameterized as multilayer perceptrons (MLP) show impressive performance in 3D shape segmentation \cite{wang2019dynamic, chen2019bae}. These methods first leverage encoders to regress a latent shape embedding, which is then decoded into a shape-conditioned 3D segmentation function via concatenation-based conditioning. 

\begin{figure*}
\begin{center}
\includegraphics[width=10.5cm]{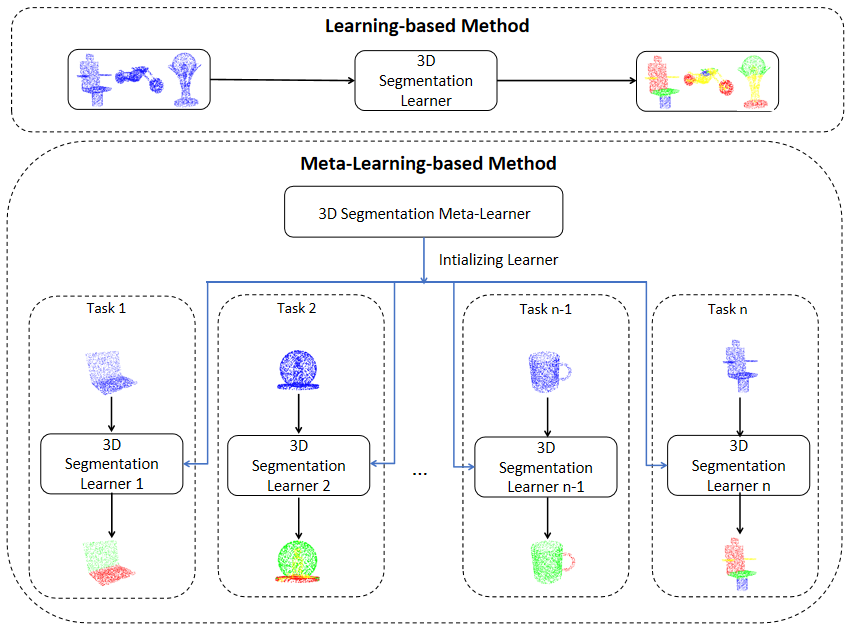}
\end{center}
\caption{Structure comparison among learning-based segmentation methods and meta-learning-based segmentation methods. Previous learning-based methods regard part segmentation as a task for the given point clouds. In comparison, we define each task as learning a unique 3D segmentation function (learner) for each given 3D object. By training over several 3D part segmentation tasks, our 3D segmentation meta-learner can dynamically update the 3D segmentation learner with optimal parameters which enables our part segmentation learner to rapidly adapt to new part segmentation tasks.}
\label{fig00}
\end{figure*}

To generate consistent part segmentation of shapes in presence of significant 3D structure variations, it is important to generalize across 3D shape segmentation functions which often requires robust learning of priors over the respective function space. To this end, existing generalization methods rely on extensive training of 3D shape segmentation functions on large-scale labeled datasets. For example, in \cite{qi2017pointnet, qi2017pointnet++}, those methods generalize their 3D segmentation ability from training a large volume of data across different categories with the hope of directly predicting the desired part labels at testing. In \cite{chen2019bae}, the authors developed the per-category 3D segmentation model in a few-shot setting, which trains a 3D segmentation model for every category of 3D objects. 

Traditional learning-based methods with a small number of training datasets can severely lead to overfitting problems and unsatisfied performance. Meta-learning \cite{andrychowicz2016learning, schmidhuber1992learning, hospedales2020meta} demonstrated its effectiveness in 2D computer vision field. As a subfield of machine learning, it introduced a novel approach to focus on task distribution other than the data distribution to faster and more efficiently adapt the learned pattern to a novel dataset with zero/few shots.

As shown in Figure \ref{fig00}, previous learning-based methods tend to learn a single 3D segmentation function (learner) for all 3D objects. Under this circumstance, the 3D segmentation learner generalizes their ability to segment a shape from extensive training on a large number of labeled data. In contrast, we proposed to formalize the learning of a 3D shape segmentation function space as a meta-learning problem, aiming to predict a 3D segmentation model that can be quickly adapted to new shapes with no or limited training data. More specifically, as shown in Figure \ref{fig00}, we define each task as learning a unique 3D segmentation function (learner) for each given 3D object. Besides, we design an auxiliary deep neural network as a meta-learner that can predict the prior over the respective 3D segmentation function space. As shown in Figure \ref{fig00}, the meta-learner is responsible for providing the optimal initialization of a 3D segmentation learner that is specialized to a new task (i.e. segmenting a new 3D object) via a few steps of gradient descent. In general, our meta-learning-based approach gains competitive advantages over existing generalization methods in that our Meta-3DSeg can uniquely parameterize the 3D segmentation function for each shape to provide an instance-wise part segmentation.

\begin{figure*}
\begin{center}
\includegraphics[width=16cm]{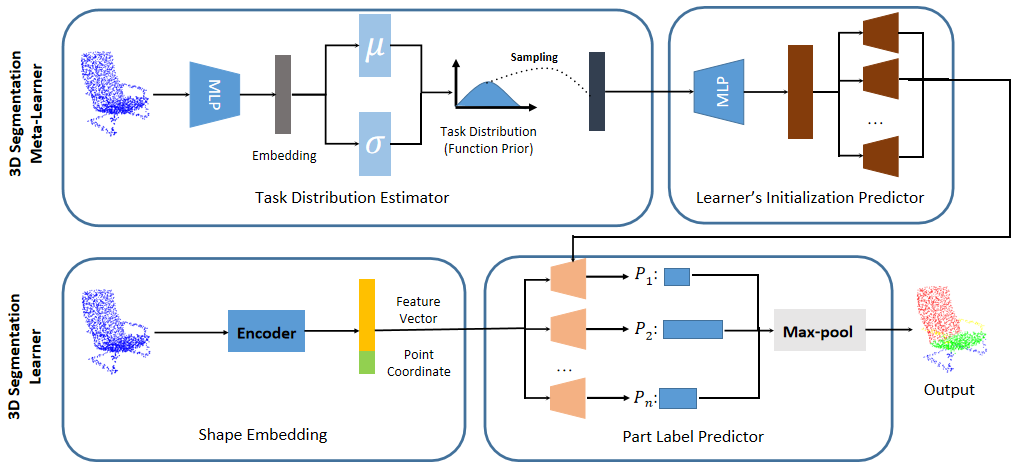}
\end{center}
\caption{Our pipeline. The proposed paradigm includes two main parts: 3D segmentation learner and 3D segmentation meta-learner. $Pn$ denotes the n-th part probability score. }
\label{main}
\end{figure*}

Our proposed Meta-3DSeg includes two modules as shown in Figure \ref{main}: 3D segmentation learner and 3D segmentation meta-learner. The 3D segmentation meta-learner includes two parts. The first part is a variational encoder that maps shape representations to a distribution over 3D segmentation tasks, which gives the priors over the 3D segmentation function space and the second part is to sample from the task distribution to predict optimal initialization for the parameters of the 3D shape segmentation function. The 3D segmentation learner is a multi-layer perceptron (MLP) conditioned on a 3D shape embedding to represent shape segmentation. Our contributions are listed as below: 1) to the best of our knowledge, it is the first time to formalize learning of a 3D shape segmentation function space as a meta-learning problem. With this, a 3D segmentation model can be quickly adapted to new shapes with no or limited labeled training data.
2) we propose a novel variational encoder that maps shape representations to a distribution over 3D segmentation tasks, contributing to robust learning of priors over the 3D segmentation function space. 3) our Meta-3DSeg can uniquely parameterize the 3D segmentation function for each shape to provide an instance-wise part segmentation. 4) we demonstrated superior 3D segmentation performance over other state-of-the-art supervised/unsupervised methods on various datasets even without training on any labeled data.

%





\section{Related Works}
\subsection{3D shape segmentation}
3D point cloud part segmentation is an important 3D computer vision task that aims to predict the part-specific label for each point of a point cloud. As we all know, traditional architectures such as voxels and image grids have regular data formats. However, the representation of point clouds is unstructured and unordered which makes the learning of point-wise labels very challenging. Early methods \cite{huang2011joint,meng2013unsupervised,sidi2011unsupervised,wu2013unsupervised} were build on the effective extraction of low-level features from 3D shapes for segmentation. Descriptors such as scale-invariant heat kernel signatures (SIHKS) \cite{wu2013unsupervised}, shape diameter function (SDF) \cite{shapira2008consistent}, Gaussian curvature (GC) \cite{gal2006salient} and so on, are widely used for this task. For 3D shape segmentation, PointNet \cite{qi2017pointnet} firstly proposed an efficient way to directly learn the features from unordered 3D point sets. Following works including PointNet++ \cite{qi2017pointnet++}, SO-Net \cite{li2018so}, SplatNet \cite{su2018splatnet}, PointCNN \cite{li2018pointcnn}, KPConv \cite{thomas2019kpconv} and so on, further improved the PointNet for 3D point cloud segmentation from various aspects such as points density, points sampling, local feature extraction and so on. Wang et al. proposed DGCNN \cite{wang2019dynamic} which uses a module called the EdgeConv that can capture local structures. The BAE-Net \cite{chen2019bae} successfully introduced a branched autoencoder network for the co-segmentation of 3D shapes. Although these previous researches provided successful attempts to design a good 3D segmentation learner for this task, large amounts of training data are required since they mostly directly predict 3D shape segmentation in the training process. In real-world scenarios, training data is expensive and hard to acquire which limits the use of those strong supervised methods. In comparison, our paper focuses on how to build a 3D segmentation meta-learner that can provide optimal initialization for the segmentation learner's parameters based on the priors over the 3D segmentation function space. 

\subsection{Meta Learning Network}
Traditional learning-based methods with a small number of training datasets can severely lead to overfitting problems and unsatisfied performance. Meta-learning \cite{andrychowicz2016learning, schmidhuber1992learning} demonstrated its effectiveness in 2D computer vision field. As a subfield of machine learning, it introduced a novel approach to focus on the task distribution other than the data distribution to faster and more efficiently adapt the learned pattern to a novel dataset with zero/few shots. Parameters prediction \cite{finn2017model, finn2018probabilistic} is one of the strategies in meta-learning, which refers to a network that is trained to provide optimal initialization of another network via gradient descent. MAML \cite{finn2017model} proposed by Finn et al. is a pioneer work in meta-learning that starts multiple tasks at the same time and learns a common best base model by different tasks. Santoro et al. \cite{santoro2016meta} introduced the model MANN to use an explicit storage buffer to rapidly learn the new task information. Meta-SDF \cite{sitzmann2020metasdf} was further proposed by Sitzmann et al. which leverages gradient-based meta-learning for the learning of neural implicit function spaces. In this paper, we first propose a meta-learning strategy that uses a 3D segmentation meta-learner to learn the general information over a variety of part segmentation tasks so that it can provide optimal initialization of the 3D segmentation learner and enable the 3D segmentation learner's network to rapidly adapt to the new part segmentation tasks.

\section{Methods}\label{headings}


\subsection{Problem Statement}\label{set1}

We define our meta-learning formulation at first. Giving a 3D point cloud \(P\) from the dataset \(\mathbf{D}\). We assume the existence of a parametric function \(g_{\theta}(P) = \phi\) using a neural network structure, where \(\phi\) is the part label estimation function which is able to predict the part label of each point in a point cloud. The 3D segmentation learner \(g\) includes two sets of parameters: \(\theta_m\) and \(\theta_l\). \(\theta_m\) is predicted by another parametric function \(f_\sigma\) which is called 3D segmentation meta-learner and \(\theta_l\) is learned during the fine-tune process. We aim to optimize the weights of the 3D segmentation meta-learner using a stochastic gradient descent-based algorithm. For a given dataset \(\mathbf{D}\), we optimize the \(\sigma\) and \(\theta_l\) as follows:

\begin{equation}
\boldsymbol{\sigma^{*}}, \boldsymbol{\theta_l^{*}} = \mathop{\arg\min}_{\sigma, \theta_l} \left[\mathbb{E}_{(P) \sim \mathbf{D}}\left[\mathcal{L}\left(g_{(f_{\sigma}(P), \theta_l)}(P)\right)\right]\right]
\end{equation}
where $\mathcal{L}$ represents the pre-defined loss function.


\subsection{3D segmentation learner}\label{set2}

\begin{figure*}
\centering
\includegraphics[width=12cm]{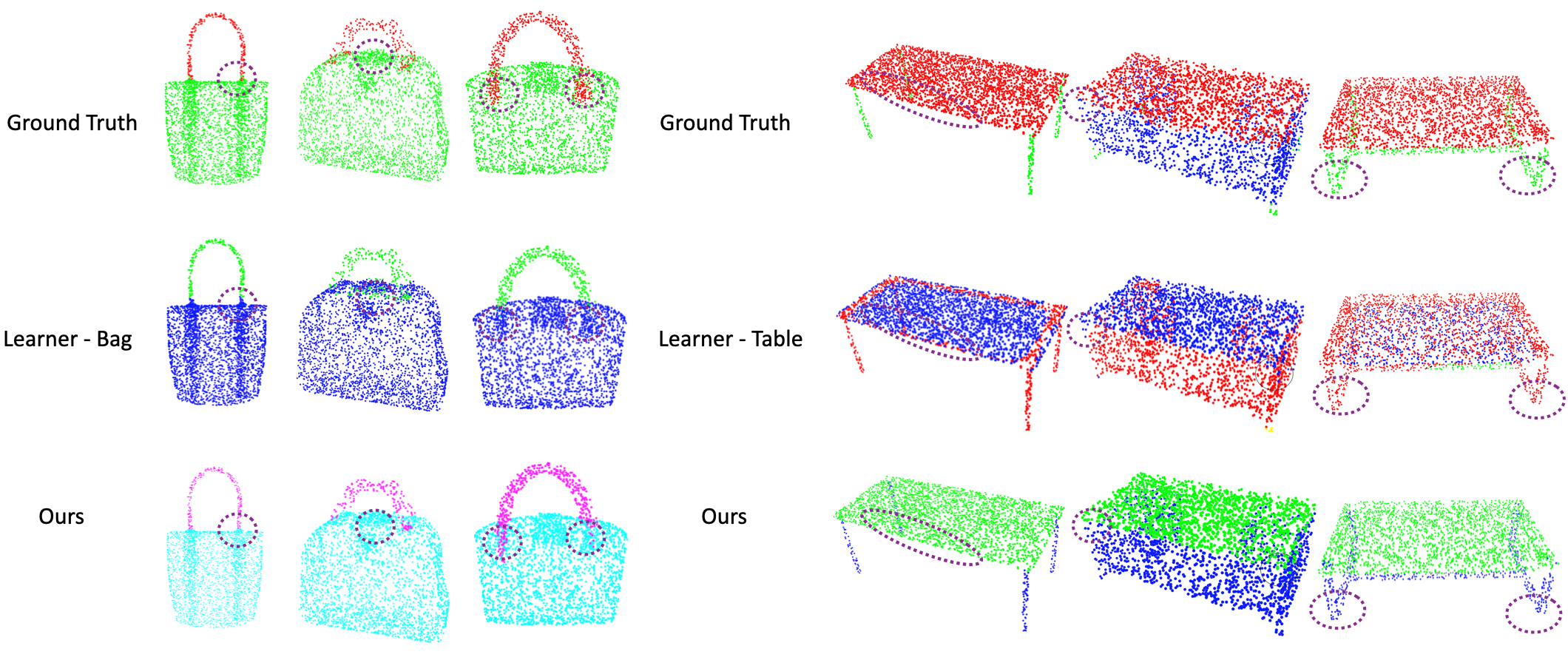}
\caption{ Ablation study. Randomly selected qualitative results from various versions of our proposed method. }
\label{fig:4}
\end{figure*}



For the input point clouds, we utilize a MLP-based neural network to extract shape features and capture geometric information \cite{chen2019bae}. Formally, let $P$ denotes the input point clouds and $f_x$  denotes the feature of $x$, $\forall x\in P$. We define the MLP-based encoding network $f_1: \mathbb{R}^{3} \to \mathbb{R}^m$ function that maps the input point clouds to the corresponding features, where $m$ denotes the dimension of features. We further concatenate the learned feature $f_x$ with the coordinates $x$ as the shape embedding $[f_x, x] \in \mathbb{R}^{m+3}$.
 
For the part label predictor, our key idea is to use a set of MLPs to predict the probability scores of all parts for each point on the 3D shape. In other words, if two points belong to the same part, they have a stronger response with a higher probability score on the same MLP. With this design, our 3D part segmentation module first takes as input the concatenation of the point coordinate and its corresponding shape features. Then, our 3D part segmentation module performs a regression that maps the shape embedding from $\mathbb{R}^{(m + 3)} \to \mathbb{R}^n$ outputting the predicted probability scores of all parts for each point, where n refer to the pre-defined number of 3D parts. 
Formally, we define a set of MLP-based functions $\{h_k\}_{k=1,...,n}$ with a softmax activation function to predict the probability score $s_k$ that indicates the likelihood that the given point belongs to a particular part: 
\begin{equation}
\begin{split}
s_k= Softmax(h_k([f_x, x]))_{x \in P}
\end{split}
\end{equation}
where $s_k$ denotes the probability score of the k-th part.

After we have the probability scores of all parts that the given point belongs to, we use a max-pooling function to select the highest probability score among all the pre-defined parts and output the predicted part label $o$ of the input point.
Note that our model is trained with a self-supervised reconstruction loss \cite{chen2019bae}. While predicting the part label, our model is able to naturally estimate the inside-outside status for each 3D point in the shape.

\subsection{3D Segmentation meta-learner} \label{set4}

To enable the 3D segmentation meta-learner to learn the function priors over the 3D segmentation function space, we introduce a variational auto-encoder (VAE) network \cite{kingma2013auto}. More specifically, we use a multi-layer MLP-based function $f_1: \mathbb{R}^m+3 \to \mathbb{R}^v$ to learn the mean and variance of the task distribution. We denote $(\bold{\mu_{m}}, \bold{\sigma_{m}})$ as the mean and the standard deviation of 3D segmentation function space:
\begin{equation}
\begin{split}
\bold{\mu_{m}}, \bold{\sigma_{m}}=f_1([f_v, x])_{x \in P}
\end{split}
\end{equation}
Therefore, we are able to sample $I_m \sim N(\mu_{m},\sigma_{m})$.
Then, we use the multi-layer MLP based architecture $f_2: \mathbb{R}^v \to \mathbb{R}^w$, where $w$ indicates the dimension of all the weights included in $\{\theta_m\}$. The meta-learned weights in the 3D segmentation learner are predicted from the following:

\begin{equation}
\begin{split}
\bold{\theta_m}=f_2(I_m)
\end{split}
\end{equation}

Therefore, our 3D part segmentation learner can be rapidly adapted by adding the meta-learned weights of $\theta_m$ from the 3D segmentation meta-learner together with the learned weights of $\theta_l$.

\section{Experiments}

\subsection{Dataset Preparation} \label{exp1}
We evaluate our proposed method for 3D point cloud part segmentation on ShapeNet part dataset \cite{yi2016scalable}. The ShapeNet dataset is an open-source collection, which consists of 16,881 shapes from 16 object categories. Each object category has 2 to 5 part labels(50 in total). To ensure a fair comparison, we follow the official split to preprocess our dataset \cite{wu20153d}. For the evaluation of point cloud part segmentation performance, we use mean IoU on points as our metric. 

\subsection{Ablation Study} \label{exp3}

\noindent{\textbf{Experiment setting:}}
In this experiment, we conduct a series of experiments to examine and verify the effects under different model initialization and settings. In the weight-setting-A, without our meta-learning strategy, the network parameters of the 3D segmentation learner are directly learned during the training process using shapes in the chair category. In the weight-setting-B, the network parameters of the 3D segmentation learner are initialized by the 3D segmentation meta-learner and then fine-tuned using shapes of the chair category. Specifically, we first sample multiple tasks using all shapes in the original dataset of all categories except for the chair category as our meta dataset. We trained our 3D segmentation meta-learner on the meta dataset to learn the priors over the 3D segmentation function space. Then we fix our 3D segmentation meta-learner and fine-tune our 3D segmentation learner using all shapes in the chair category. Note that we also pre-trained the model with weight-setting-A on the meta dataset for a fair comparison. 

\begin{table}[b]
  \caption{ Quantitative result. Ablation study on the ShapeNet dataset.}
  \label{table1}
  \centering
  \small
  \begin{tabular}{lll}
    \hline                  
    Models     & IoU\%& Acc\%\\
    \hline
    Learner with Weight-Setting-A&83.7&88.7 \\
    Learner with Weight-Setting-B&88.4&92.9 \\
    Learner with Weight-Setting-C&90.2&94.4\\
    \hline
  \end{tabular}
  \vspace{-5mm}
\end{table}

\begin{table*}
  \caption{Quantitative result. Comparison with other supervised methods}
  \label{table3}
  \centering
  \small
  \begin{tabular}{l|l|l|llll|l}
    \hline                  
    Categories     & Ours &BAE-NET\cite{chen2019bae} & WPS-NET\cite{wang2020few} & PointNet\cite{qi2017pointnet} &PointNet++\cite{qi2017pointnet} &PointConv\cite{{Wu_2019_CVPR}} & PointNet (all)\\
    \hline
    Supervision     & N &N & 10 & 10 &10 &10& All\\
    \hline
    Airplane$_4$ & \textbf{62.3} & 61.1 & 67.3 & 63.3 & 62.3 &  65.1 & 83.4 \\
    Airplane$_3$ & \textbf{82.4} & 80.4 & - - - & - - - & - - - & - - - & - - -\\
    Bag & 81.3 & \textbf{82.5} & 74.4 & 64.9 &67.4 & 68.2   & 78.7  \\
    Cap & 86.4 & \textbf{87.3} & 86.3 & 75.2 & 80.0 & 80.7   & 82.5  \\
    Chair$_4$ &\textbf{68.8}  & 65.5 & 83.4 & 73.8 & 61.6 & 66.1 & 89.6\\
    Chair$_3$ & \textbf{90.2} & 86.6 & - - - & - - - & - - - & - - - & - - - \\
    Mug & \textbf{94.1} & 93.4 & 90.9 & 80.9 & 83.1 & 86.0 & 93.0 \\
    Table\_3 &  \textbf{79.7}&  78.7& 74.2 & 72.2 & 72.2 & 72.5 & 80.6\\
    Table\_2 & \textbf{89.9} & 87.0 & - - - & - - - & - - - & - - - & - - - \\
    \hline
  \end{tabular}
\end{table*}

In the weight-setting-C, with our meta-learning strategy, instead of directly using the MLPs to map the input shape to the function prior, we add a variational auto-encoder(VAE) to learn the task distribution of the input 3D shape, which can be further sampled as the function prior over 3D segmentation function space for the learner predictor to estimate the optimal weights of the 3D segmentation learner. 

We further verify the efficiency of our model under different model initialization and settings in Figure \ref{fig:4}.  As for the learner for a particular category, without a meta-learning strategy, the network parameters of the 3D segmentation learner are learned during the training process. As for our model, we leverage the proposed 3D segmentation meta-learner to learn the prior over the 3D segmentation function space and provide the optimal parameters of the 3D segmentation learner. 


\noindent{\textbf{Results:}}
By comparing the results of the first two rows in Table \ref{table1}, the performance gain indicates that our meta-learning method can learn the general information over several part segmentation tasks and benefit the part segmentation learner from good meta-learned initialized weights. Moreover, as indicated in row 2 and 3, the experimental results of the weight-setting-C are better than the weight-setting-B, which indicates that the task distribution generated by the VAE in the 3D segmentation meta-learner can gather the prior over the 3D segmentation function space and further sampled for the learner predictor to predict the optimal parameters of the 3D segmentation learner. This indicates that our meta-learning strategy can learn the general information over multiple part segmentation tasks and benefit the part segmentation learner from good meta-learned network parameters. Besides, as illustrated from the qualitative results shown in Figure \ref{fig:4}, our method is more accurate in estimating part-specific labels of point clouds. 

\begin{figure*}
\centering
\includegraphics[width=12cm]{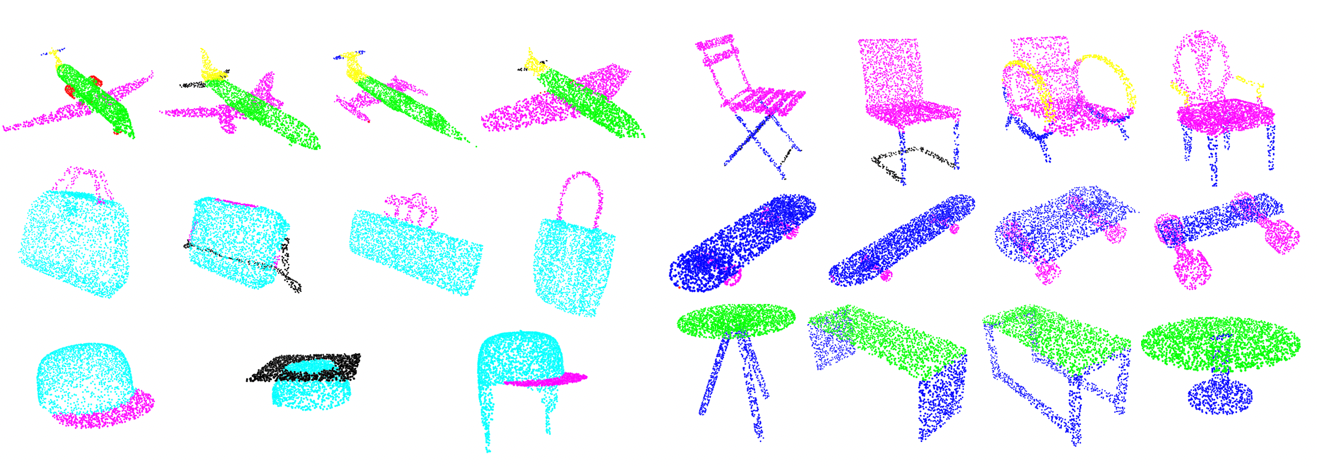}
\caption{ Qualitative results. Randomly selected qualitative results of point clouds part segmentation on multiple categories of the ShapeNet dataset.}
\label{fig:5}
\end{figure*}


\vspace{-3mm}
\subsection{Comparative Study} \label{exp5}
\vspace{-1mm}

\noindent{\textbf{Experiment setting:}} In this experiment, we evaluate the general segmentation performance extended to more shape categories and compare the performance with other state-of-the-art approaches. 
The second row shows the supervision state for each method where N denotes that the model is trained in an unsupervised manner without any ground truth part labels, 10 denotes that the model is trained using 10 randomly chosen samples in a specific category and all denotes that the model is fully supervised trained with all shapes in that category. Note that to enhance its performance, BAE-NET is trained category-by-category which means there will be a category-specific BAE-NET model for each category after training. In comparison, our model is trained across different categories which enables the robust learning of priors over the respective function space. Furthermore, we use the subscript to specify the number of classes since various algorithms probably would generate different segmentation results. For instance, our model and BAE-NET would occasionally segment the airplane object into three parts rather than four due to the incapability of splitting wings and engines. 

\noindent{\textbf{Results:}} Table \ref{table3} presents the per-category IoU for each model we trained on. The data indicates that our model achieves much better performance on most shape categories contrasted to all supervised methods under the sample training process. 
The inherent training difficulty of our Meta-3DSeg model is the hardest since it is purely unsupervised and trained across all categories. Nevertheless, the training result is still promising and improved from previous methods, which justifies the effectiveness of our meta-learning design. Compared with the PointNet result using all training samples, our approach only lags behind by a reasonable margin. 
Figure \ref{fig:5} shows some randomly selected qualitative results of our Meta-3DSeg on novel categories of the ShapeNet dataset.

\vspace{-3mm}
\section{Conclusion}
\vspace{-3mm}
In this paper, we introduce a novel meta-learning strategy to our research community for 3D point cloud part segmentation. Compared with previous learning-based methods, our Meta-3DSeg can learn the distribution of tasks instead of the distribution of data. With the prior over the respective 3D segmentation function space over multiple similar tasks, our Meta-3DSeg is capable to rapidly adapt and have good generalization performance on new tasks. Note that, our proposed meta-learning strategy can hopefully benefit our 3D community by providing an efficient and effective learning algorithm to train the mainstream 3D part segmentation learners. To the best of our knowledge, our method firstly leveraged a novel meta-learning strategy for this task and we experimentally verified the effectiveness of our model and achieved superior unsupervised part segmentation results on the ShapeNet 3D point cloud part segmentation dataset. 

\bibliographystyle{IEEEbib}
\bibliography{egbib}

\end{document}